\documentclass[journal]{IEEEtran}
\usepackage{makecell}
\usepackage[colorlinks]{hyperref}
\usepackage{multirow}
\usepackage{amssymb}
\usepackage{amsmath}
\usepackage{comment}
\usepackage{float}
\usepackage{subfigure}
\usepackage{subcaption}
\ifCLASSINFOpdf
\else
   \usepackage[dvips]{graphicx}
\fi
\usepackage{url}

\usepackage{graphicx}

\begin{document}

\title{ResEmoteNet: Bridging Accuracy and Loss Reduction in Facial Emotion Recognition}

\author{Arnab Kumar Roy, Hemant Kumar Kathania, Adhitiya Sharma, Abhishek Dey and Md. Sarfaraj Alam Ansari \IEEEmembership{}

\thanks{Arnab Kumar Roy is with Sikkim Manipal Institute of Technology (SMIT), Sikkim, India - 737136 (e-mail: arnab\_202000152@smit.smu.edu.in),  Hemant Kumar Kathania, Adhitiya Sharma, and Md. Sarfaraj Alam Ansari are with National Institute of Technology Sikkim, India - 737139 (e-mail: hemant.ece@nitsikkim.ac.in, b180078@nitsikkim.ac.in, sarfaraj@nitsikkim.ac.in),  Abhishek Dey is with Bay Area Advanced Analytics India (P) Ltd, A Kaliber.AI company, Guwahati, India - 781039 (e-mail: abhishek@kaliberlabs.com) }}

\maketitle

\begin{abstract}

The human face is a silent communicator, expressing emotions and thoughts through it's facial expressions. With the advancements in computer vision in recent years, facial emotion recognition technology has made significant strides, enabling machines to decode the intricacies of facial cues. In this work, we propose ResEmoteNet, a novel deep learning architecture for facial emotion recognition designed with the combination of Convolutional, Squeeze-Excitation (SE) and Residual Networks. The inclusion of SE block selectively focuses on the important features of the human face, enhances the feature representation and suppresses the less relevant ones. This helps in reducing the loss and enhancing the overall model performance. We also integrate the SE block with three residual blocks that help in learning more complex representation of the data through deeper layers. We evaluated ResEmoteNet on four open-source databases: FER2013, RAF-DB, AffectNet-7 and ExpW, achieving accuracies of 79.79\%, 94.76\%, 72.39\% and 75.67\% respectively. The proposed network outperforms state-of-the-art models across all four databases. The source code for ResEmoteNet is available at \url{https://github.com/ArnabKumarRoy02/ResEmoteNet}.
\end{abstract}


\IEEEpeerreviewmaketitle

\vspace{-1.5em}
\section{Introduction}
Facial Emotion Recognition (FER) is a specialized task within Image Recognition, focusing on identifying emotions from facial images or videos. Facial emotions change with subtle movements of facial features such as lips, teeth, skin, hair, cheekbones, nose, face shape, eyebrows, eyes, jawline, and mouth, making it difficult to design models that can accurately capture these intricate details. Additionally, data collection for FER is a labor-intensive process that requires significant funding and careful annotation by humans. Despite its challenges, FER is a crucial task in image recognition. Facial emotions provide valuable insights into a person's mental health, helping to identify signs of depression, anxiety, and other psychiatric disorders \cite{hamm2011automated}. Consequently, FER plays a significant role in mental health and therapy. It can also be instrumental in creating responsive and adaptive systems for human-computer interaction. In educational settings, FER can help teachers understand the emotions of their students in a classroom, allowing them to use this feedback to enhance the learning experience \cite{bosch2016detecting}.

\begin{figure*}
    \vspace{-3em}
    \centering
    \label{fig:cm-label}
    \subfigure[\label{fig:se_fig}]
    {\includegraphics[height=6.0cm, width=0.2\linewidth]{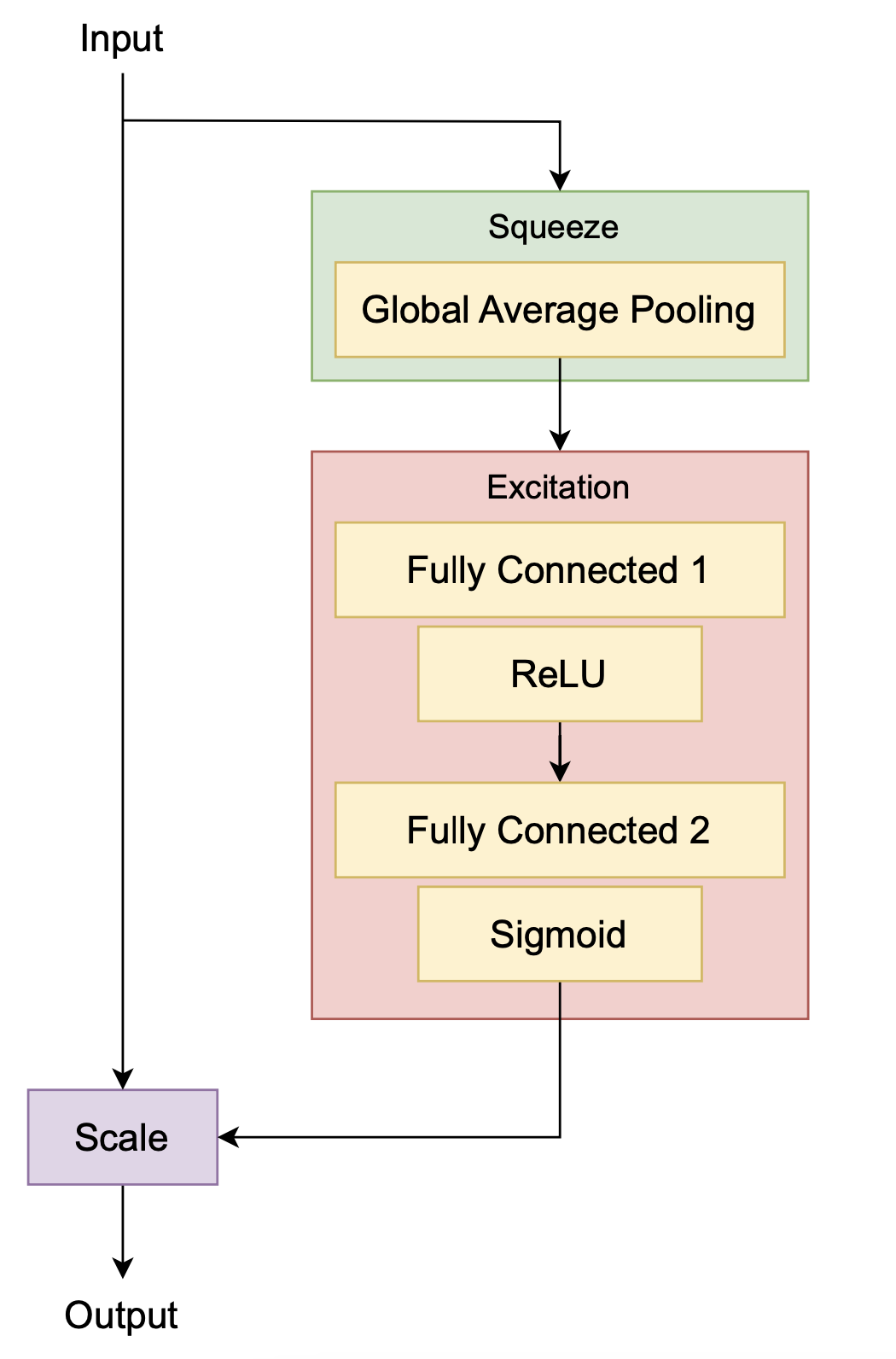}
    } \hfill
    \subfigure[\label{fig:resemotenet_fig}]
    {\includegraphics[height=6.5cm, width=0.5\linewidth]{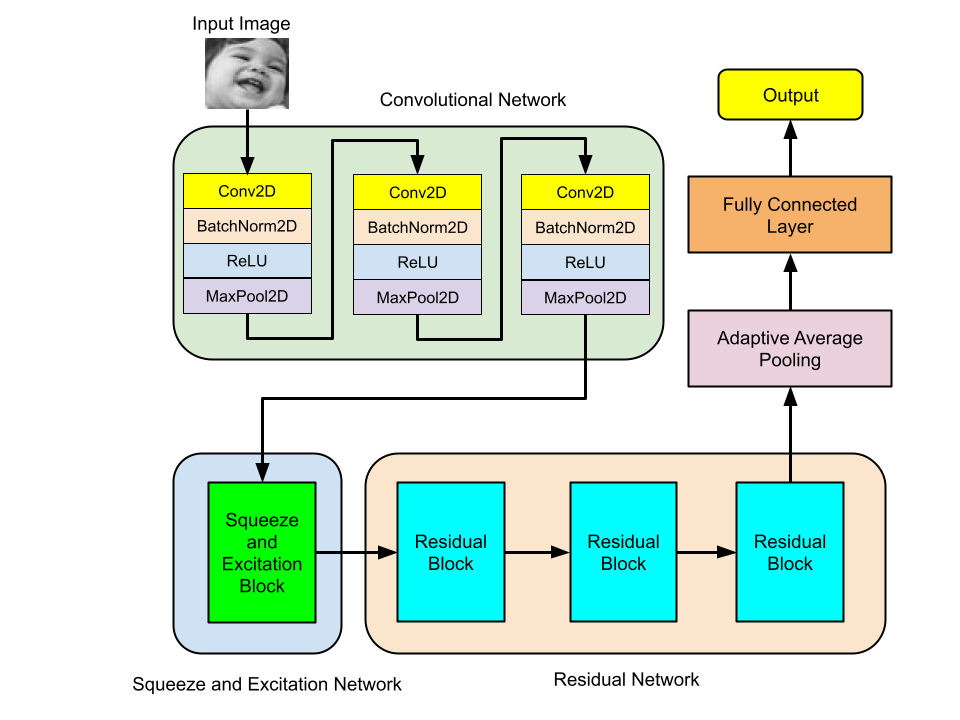}
    } \hfill
    \subfigure[\label{fig:residual_fig}]
    {\includegraphics[height=6.0cm, width=0.2\linewidth]{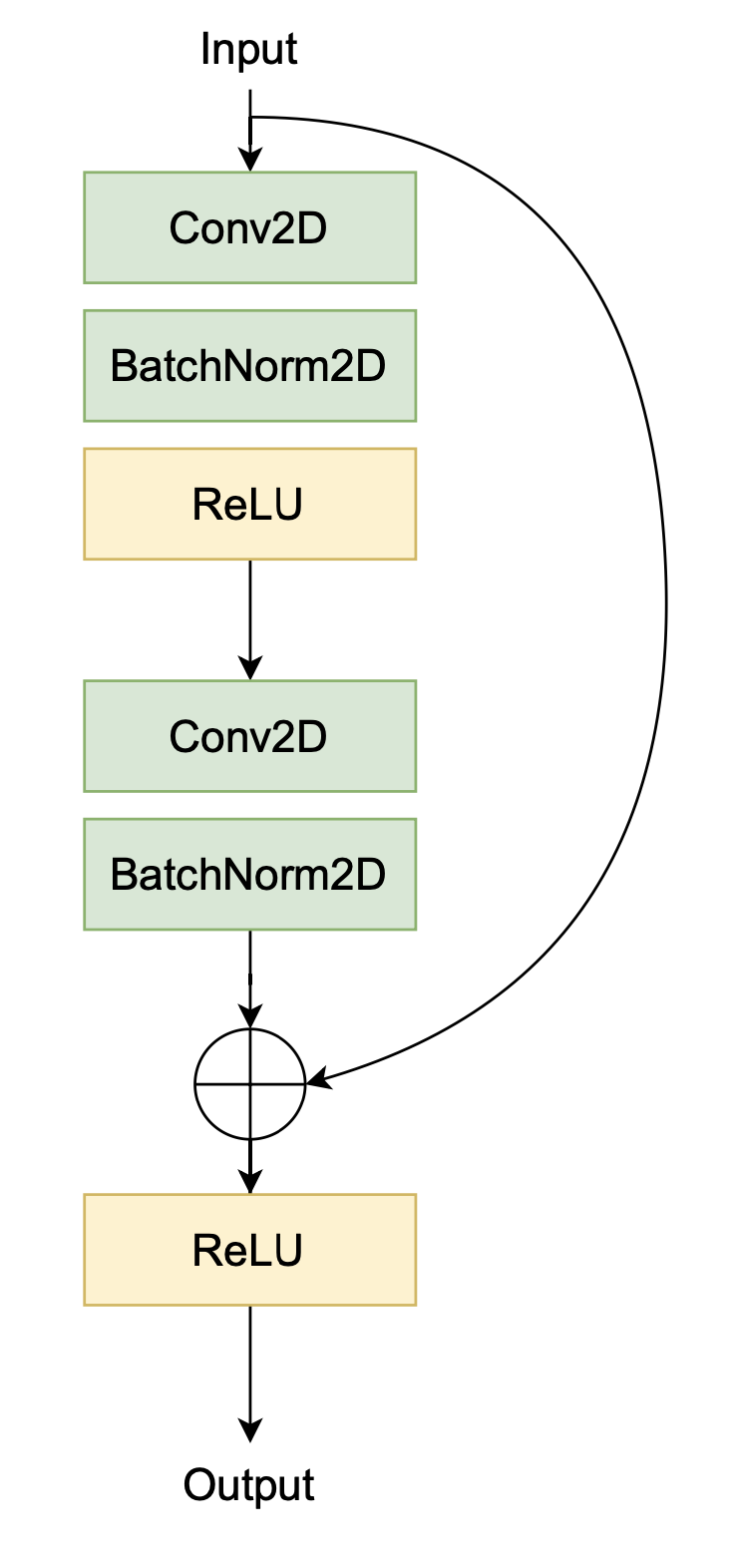}
    } \hfill
    \hfill
    \vspace{-0.5em}
    \caption{(a) Architecture of Squeeze and Excitation Block, (b) Overall Architecture of Proposed ResEmoteNet (c) Architecture of Residual Block} 
   \label{architecture_diagram}
   \vspace{-1.5em}
\end{figure*}

In recent years, FER has been dominated by deep learning systems, primarily leveraging deep Convolutional Networks such as ResNets \cite{he2016deep} and AlexNets \cite{krizhevsky2012imagenet}. Vision Transformers \cite{dosovitskiy2020image}, widely used in Natural Language Processing (NLP), have also been applied to FER, significantly enhancing performance. In \cite{zhang2023dual}, MobileFaceNet \cite{chen2018mobilefacenets} was employed as a backbone for feature extraction from images, and a Dual Direction Attention Network (DDAN) was proposed. DDAN generates attention maps from both vertical and horizontal orientations, guiding the model to focus on specific facial features and providing detailed feature representation. An attention loss mechanism was introduced to ensure that different attention heads concentrate on distinct features.

In \cite{multiheadchannel}, the Local Multi-Headed Channel (LHC) module was introduced to add channel-wise self-attention to existing CNN architectures. Pecoraro et al. incorporated LHC into ResNet34, creating LHC-Net for facial emotion classification. A similar approach is found in \cite{wen2023distract}, where ResNet is combined with spatial and channel attention mechanisms and two loss functions. In \cite{poster++}, a two-stream feature extraction method using an image backbone and facial landmark detector was proposed, utilizing cross-attention for classification. This method is more computationally efficient than \cite{zheng2023poster}.

In \cite{s2d}, Vision Transformers (ViTs) with Multi-View Complementary Prompters (MCPs) were proposed for FER, where ViT extracts facial features and MCP enhances performance by combining landmark features. An ensemble approach is introduced in \cite{residualmasking}, using the ResMaskingNetwork with other networks. In \cite{kollias2024distribution}, a multi-task learning approach integrates the EmoAffectNet and EffNet-B2 models, using a coupling loss to improve facial feature learning across datasets. In \cite{wang2024lrdif}, an attention mechanism with self-supervised learning, diffusion-based denoising, and restoration improves performance in noisy environments like under-display cameras (UDC). GroupGAN, proposed in \cite{niu2022four}, includes an extractor, generator, and two discriminators to handle weak emotions by converting them to peak emotions, improving performance. Finally, \cite{wang2024htnet} introduces HTNet for micro-expression recognition, leveraging local temporal, local, and global semantic features, demonstrating enhanced results across four publicly available datasets.

In this paper, we propose ResEmoteNet, a novel neural network architecture for facial emotion recognition. ResEmoteNet integrates Convolutional Neural Networks, Residual connections, and the Squeeze and Excitation network \cite{hu2018squeeze} to effectively capture facial emotions. We evaluated the proposed network using four open-source databases: FER2013 \cite{goodfellow2013challenges}, RAF-DB \cite{li2017reliable}, AffectNet-7 \cite{mollahosseini2017affectnet} and ExpW (Expressions in the Wild) \cite{zhang2018facial}. ResEmoteNet demonstrated state-of-the-art performance across all four databases.

\begin{figure*}[!ht]
    \centering
    \vspace{-3em}
    \includegraphics[width=320px]{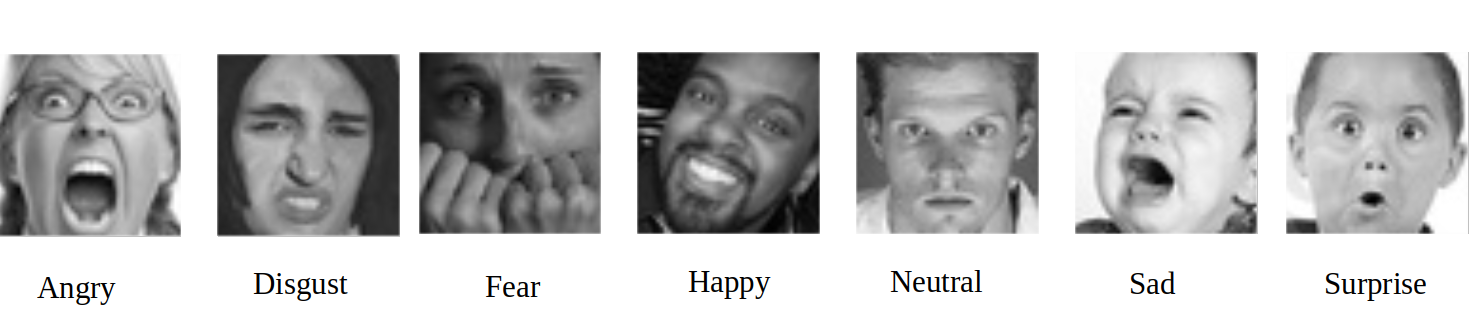}
    \caption{Representative images of the 7 emotion classes: Angry, Disgust, Fear, Happy, Neutral, Sad, and Surprise, showcasing the varied facial expressions in the datasets.}
    \label{FER-examples}
    \vspace{-1em}
\end{figure*}

\vspace{-0.2cm}
\section{Proposed Method}\label{sec:proposed method}

This section presents our proposed \textbf{ResEmoteNet} framework, with a comprehensive architecture illustrated in Fig. \ref{architecture_diagram} 
 (b). The framework integrates a simple Convolutional Neural Network (CNN) block, complemented by the Squeeze and Excitation (SE) block and reinforced by multiple Residual blocks, forming a robust and efficient network.

\vspace{-0.4cm}
\subsection{Convolutional Network}

Our architecture includes a CNN module with three convolutional layers for hierarchical feature extraction, followed by batch normalization to stabilize learning and enhance training efficiency. Max-pooling is applied after each layer to reduce spatial dimensions, lowering computational costs and introducing translational invariance for improved robustness. These layers form the foundation of our feature extraction process. Mathematically, the Convolutional Network's (CNet) feature extraction process is expressed as follows:
\begin{equation}
    X_{FE} = CNet(X)
\end{equation}
where X is the raw image sample and $X_{FE}$ is the output of the feature map from the CNet.

\vspace{-0.4cm}
\subsection{Squeeze and Excitation Network}

Squeeze and Excitation Network (SENet) is incorporated into our methodology to boost the representational power of convolutional neural networks. At the heart of SENet lies the SE block, a key component that models the relationships between convolutional channels. It performs two primary functions: Squeeze, which uses global average pooling to condense spatial data from each channel into a global descriptor, and Excitation, which employs a sigmoid-activated gating mechanism to capture channel dependencies. SENet's approach allows the network to learn a series of attention weights, highlighting the importance of each input element for the network's output. The architecture of the Squeeze and Excitation block is shown in Fig. \ref{architecture_diagram} (a).

Let $X_{FE}$ be the input for the squeeze operation which can be expressed as:
\vspace{-0.2cm}
\begin{equation}
    z_c = \frac{1}{H \times W} \sum_{i=1}^{H} \sum_{j=1}^{W} X_{FE}(c,i,j)
\end{equation}
Here, $H$ and $W$ are the height and widths of the feature maps, respectively, and $X_{FE}(c,i,j)$ denotes the activation at position $(i,j)$ in channel $c$.

The squeezed output $z$ is then processed through two fully connected layers: a dimensionality-reduction layer followed by a dimensionality-expansion layer, with a Rectified Linear Unit (ReLU) activation in between. The excitation operation can be formulated as:
\begin{equation}
    X_{S1} = ReLU(W_1z)
\end{equation}
\begin{equation}
    X_{S2} = W_2X_{S1}
\end{equation}
\begin{equation}
    s = sigmoid(X_{S2})
\end{equation}
Here, $W_1$ is the weight matrix that reduces the dimensionality.  $X_{S1}$ is the output from first dimension-reduction layer. $W_2$ is the weight matrix that expands it back to the original number of channels i.e., $c$.$X_{S2}$ is the output from first dimension-expansion layer. The per-channel modulation weights, derived from the excitation phase output $s$, are employed to adjust the original input feature maps $X_{FE}$. This scaling operation is done on each element individually and is represented by Y:
\begin{equation}
    Y = s \cdot X_{FE}
\end{equation}

\vspace{-0.8cm}
\subsection{Residual Network}
Residual Networks (ResNets) are a significant innovation in deep learning, particularly in fields that involve training extremely deep neural networks. He et al. \cite{he2016deep} introduced ResNets, which efficiently tackle the common issues of vanishing and exploding gradients in neural networks. ResNets' main innovation is the addition of the residual block, which includes a shortcut connection to skip one or more layers. Mathematically, the function in a residual block can be defined as:
\begin{equation}
    F(x) = H(x) + x
\end{equation}
In a residual block, $x$ is the input, $H(x)$ is the output from stacked layers, and $F(x)$ is the final output. Adding $x$ to $H(x)$ enables the network to learn identity mapping, ensuring deeper layers perform as well as shallow ones, addressing degradation. This allows training deeper networks than before, improving tasks like image classification and object detection on benchmark datasets like ImageNet \cite{deng2009imagenet}. Fig. \ref{architecture_diagram} (c) shows the architecture of a residual block.

\vspace{-1em}
\subsection{Adaptive Average Pooling}
Adaptive Average Pooling (AAP) is a technique that was first introduced in 2018 \cite{liu2018path}. AAP is a type of pooling layer used in CNNs that enables the aggregation of input information into a constant output size, regardless of the original input dimensions.
AAP adjusts kernel size and stride to reach a specific output size, instead of reducing spatial dimensions like traditional pooling methods. It ensures consistent output dimensions in various datasets and layers.

Considering $X_{RB}$ as the output feature map from the residual block and $\tilde{X}_{FE}$ be the output from the AAP operation, this operation can be expressed as:
\begin{equation}
    \tilde{X}_{FE} = AAP(X_{RB})
\end{equation}

$\tilde{X}_{FE}$ is finally fed to the classifier that outputs a probability distribution over the possible facial expressions.

\vspace{-0.3cm}
\begin{equation}
    P = Classifier(\tilde{X}_{FE})
\end{equation}

$P\in\mathbb{R}^{N}$, where $N$ is the number of facial emotion classes. $Classifier$ is the Linear layer that helps to classify image based on output of the network.

\begin{figure*}
    \centering
    \vspace{-2.5em}
    \label{fig:cm-label}
    \subfigure[\label{fer2013-cm}]
    {\includegraphics[width=0.228\textwidth]{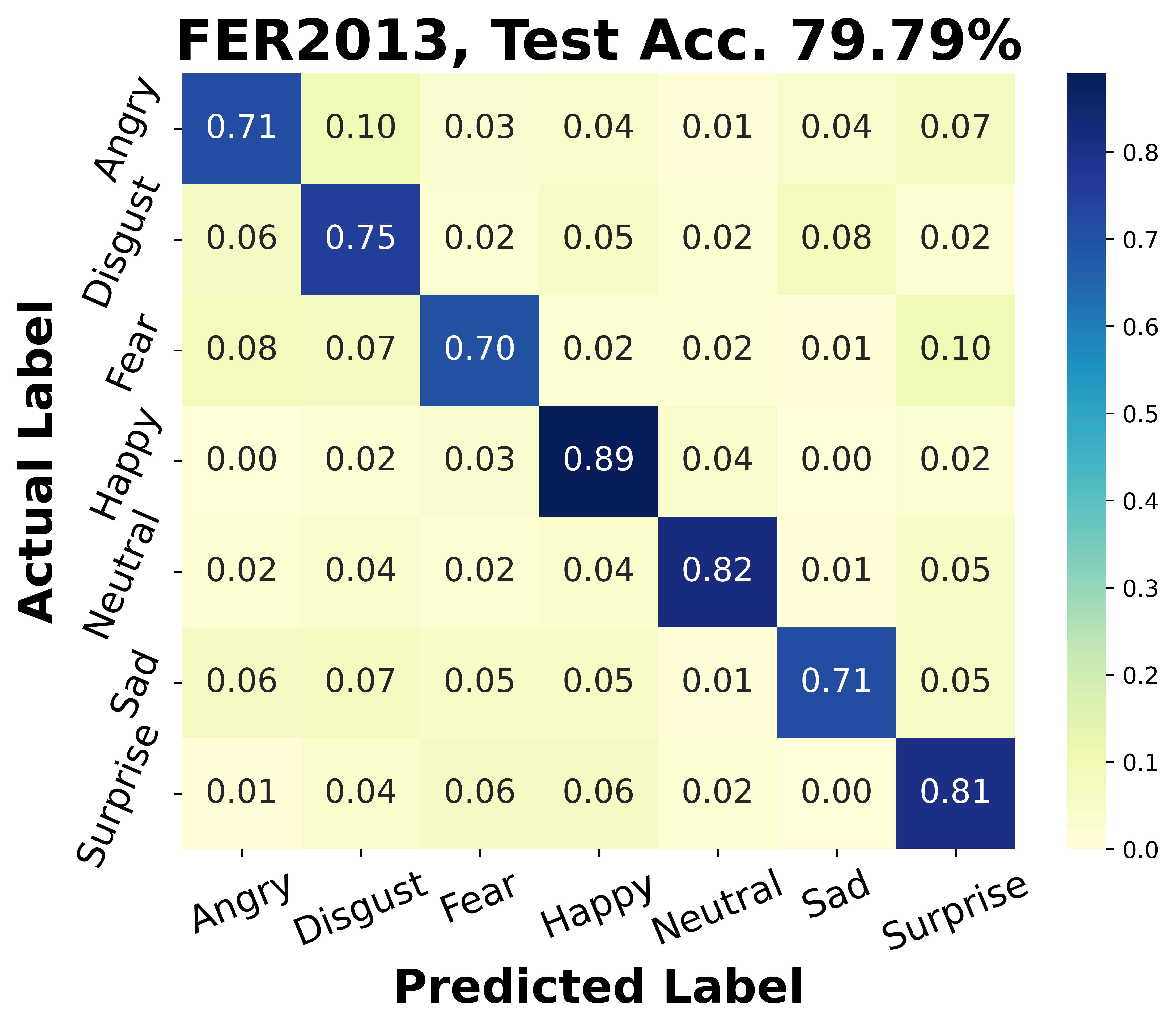}
    } \hfill
    \subfigure[\label{raf-db-cm}]
    {\includegraphics[width=0.228\textwidth]{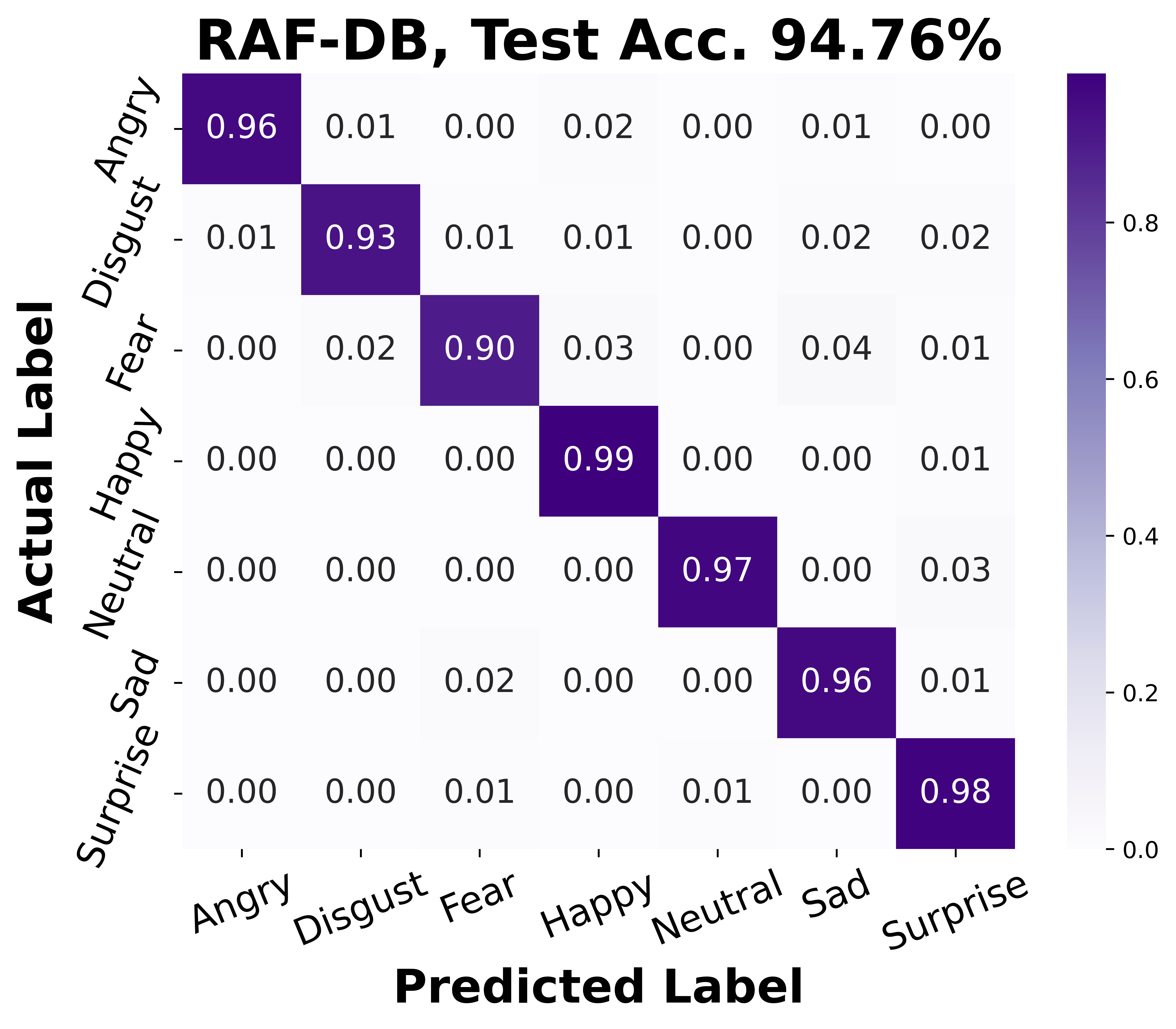}
    } \hfill
    \subfigure[\label{affectnet-cm}]
    {\includegraphics[width=0.228\textwidth]{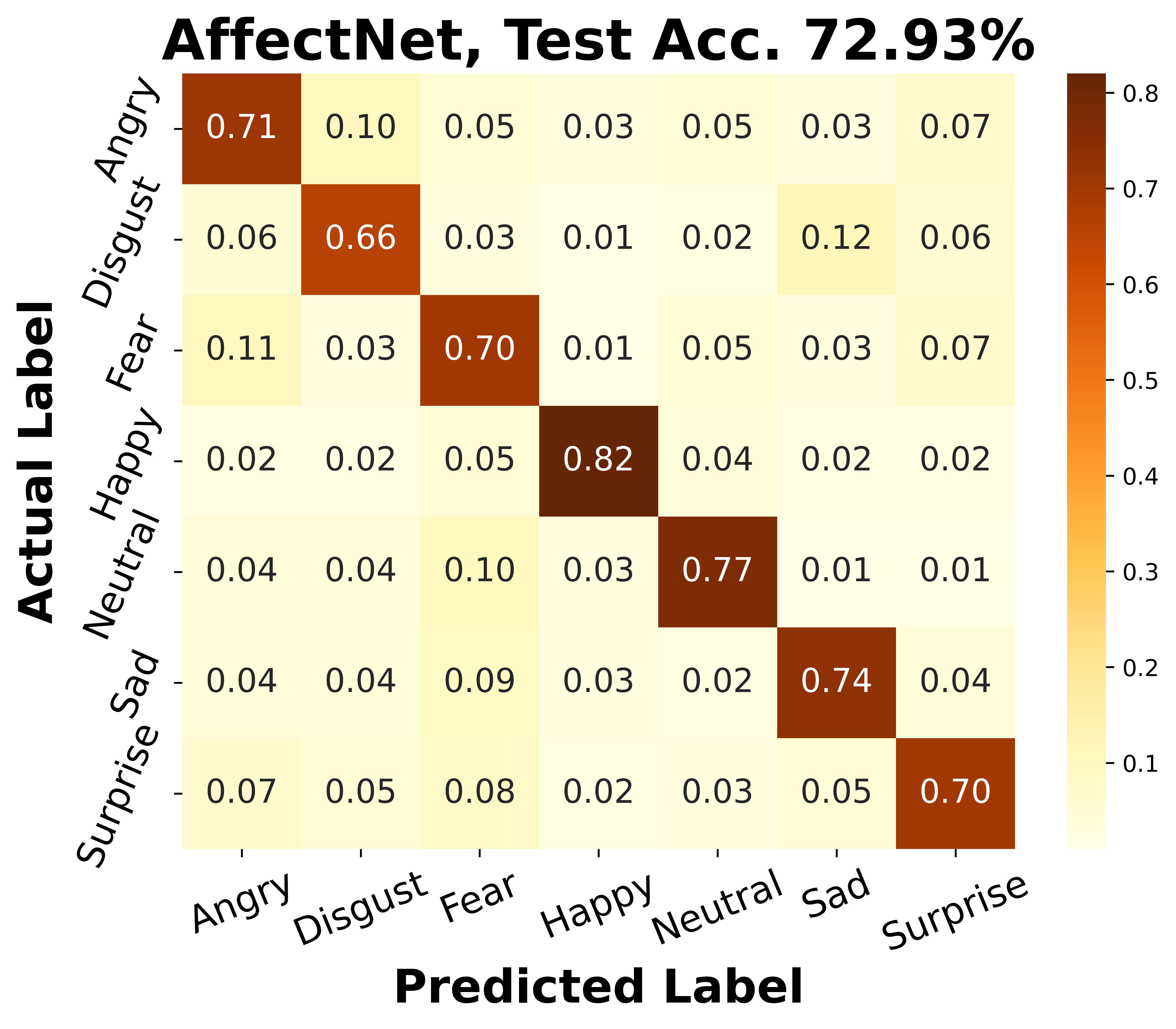}
    } \hfill
    \subfigure[\label{expw-cm}]
    {\includegraphics[width=0.228\textwidth]{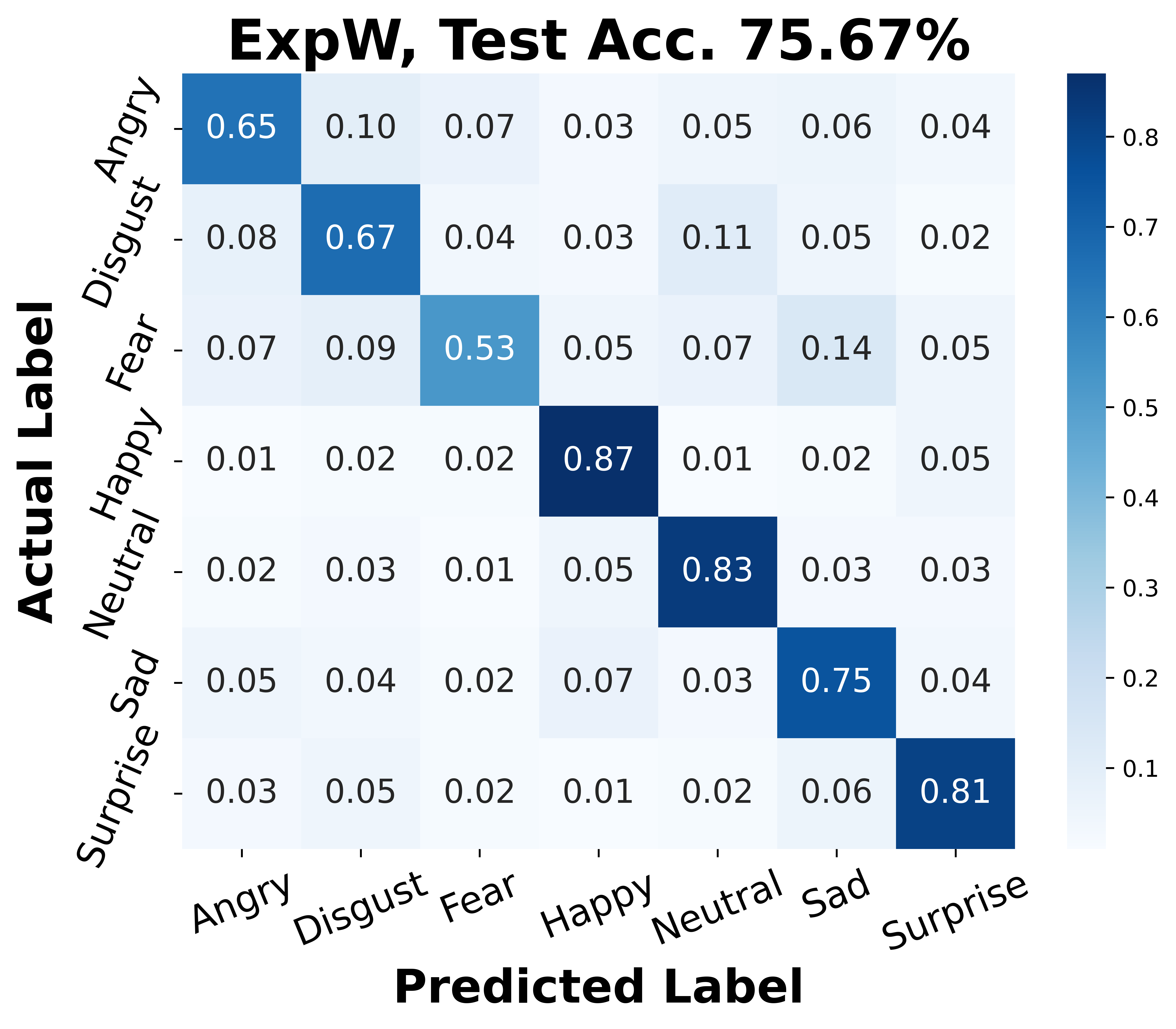}
    } \hfill
    \hfill
    \vspace{-0.5em}
    \caption{Confusion matrices of ResEmoteNet across four databases: (a) for FER2013, (b) for RAF-DB (c) for AffectNet-7 and (d) for ExpW.} 
    \label{fig:confusion_matrices}
    \vspace{-1.7em}
\end{figure*}

\vspace{-0.5cm}
\section{Dataset and Experimental details}
\label{sec:experiments}

\vspace{-0.5em}
\subsection{Dataset}
\vspace{-0.3em}

In this sub-section, we provide a brief overview of the datasets used in this study. We conducted our experimental studies on four popular Facial Emotion Recognition datasets namely \textbf{FER2013 \cite{goodfellow2013challenges}, RAF-DB \cite{li2017reliable}}, \textbf{AffectNet-7 \cite{mollahosseini2017affectnet}} and \textbf{ExpW (Expressions in the Wild)} \cite{zhang2018facial}. A comparison of the four datasets is presented in Table~\ref{data-compare}, detailing their number of channels, image sizes, number of samples and number of classes. Our facial emotion recognition task involves identifying seven fundamental emotions: Angry, Disgust, Fear, Happy, Neutral, Sad and Surprise. Visual examples of each emotion are displayed in Fig. \ref{FER-examples}. To ensure a comprehensive analysis, we have also included a class-wise breakdown of the train-test distribution for each dataset in Table \ref{classwise-data-stats}.

\vspace{-0.5em}
\renewcommand{\arraystretch}{1.1}
\begin{table}[!ht]
    \centering
    \caption{Dataset comparison}
    \scalebox{0.95}{
    \begin{tabular}{|c|c|c|c|c|c|}
        \hline
            \textbf{Characteristics} & \textbf{FER2013} & \textbf{RAF-DB} & \textbf{AffectNet-7} & {\textbf{ExpW}}\\
        \hline
        \hline
        No. of channels & 1 & 3 & 3 & 3 \\
        \hline
        Image size & ${48 \times 48}$ & ${100 \times 100}$ & ${224 \times 224}$ & ${224 \times 224}$\\
        \hline
        Total Samples & 35,887 & 15,339 & 287,401 & 91,793 \\
        \hline
  
        No. of classes & 7 & 7 & 7 & 7\\
        \hline
    \end{tabular}}
    \label{data-compare}
\end{table}

\vspace{-0.5em}
\renewcommand{\arraystretch}{1.1}
\begin{table}[!ht]
\caption{Class wise data distribution across four datasets: FER2013, RAF-DB, AffectNet-7 and ExpW}
\scalebox{0.87}{
\begin{tabular}{|c|cc|cc|cc|cc|}
\hline
\multirow{2}{*}{\textbf{Class}} & \multicolumn{2}{c|}{\textbf{FER2013}}              & \multicolumn{2}{c|}{\textbf{RAF-DB}} & \multicolumn{2}{c|}{\textbf{AffectNet-7}} & \multicolumn{2}{c|}{\textbf{ExpW}}              \\ \cline{2-9} 
                                      & \multicolumn{1}{c|}{\textbf{Train}} & \textbf{Test} & \multicolumn{1}{c|}{\textbf{Train}} & \textbf{Test} & \multicolumn{1}{c|}{\textbf{Train}}  & \textbf{Test} & \multicolumn{1}{c|}{\textbf{Train}}  & \textbf{Test} \\ \hline
Angry                                 & \multicolumn{1}{c|}{3995}           & 491           & \multicolumn{1}{c|}{705}            & 162           & \multicolumn{1}{c|}{24882}           & 500  &     \multicolumn{1}{c|}{2569} & 364     \\ \hline
Disgust                               & \multicolumn{1}{c|}{436}            & 416           & \multicolumn{1}{c|}{717}            & 160           & \multicolumn{1}{c|}{3803}            & 500   &     \multicolumn{1}{c|}{2796} & 396          \\ \hline
Fear                                  & \multicolumn{1}{c|}{4097}           & 626           & \multicolumn{1}{c|}{281}            & 74            & \multicolumn{1}{c|}{6378}            & 500    &     \multicolumn{1}{c|}{761} & 108         \\ \hline
Happy                                 & \multicolumn{1}{c|}{7215}           & 594           & \multicolumn{1}{c|}{4772}           & 1185          & \multicolumn{1}{c|}{134415}          & 500      &     \multicolumn{1}{c|}{21375} & 3024       \\ \hline
Neutral                               & \multicolumn{1}{c|}{4965}           & 528           & \multicolumn{1}{c|}{2524}           & 680           & \multicolumn{1}{c|}{74874}           & 500      &     \multicolumn{1}{c|}{24418} & 3454       \\ \hline
Sad                                   & \multicolumn{1}{c|}{4830}           & 879           & \multicolumn{1}{c|}{1982}           & 478           & \multicolumn{1}{c|}{25459}           & 500      &     \multicolumn{1}{c|}{7391} & 1046       \\ \hline
Surprise                              & \multicolumn{1}{c|}{3171}           & 55            & \multicolumn{1}{c|}{1290}           & 329           & \multicolumn{1}{c|}{14090}           & 500      &     \multicolumn{1}{c|}{4943} & 699       \\ \hline
\textbf{Total}               & \multicolumn{1}{c|}{\textbf{28709}} & \textbf{3589} & \multicolumn{1}{c|}{\textbf{12271}} & \textbf{3068} & \multicolumn{1}{c|}{\textbf{283901}} & \textbf{3500}  & \multicolumn{1}{c|}{\textbf{64253}} & \textbf{9091}\\ \hline
\end{tabular}}
\label{classwise-data-stats}
\vspace{-1.5em}
\end{table}

\subsection{Experimental Details}

In this study, we employed a consistent experimental setup across four facial emotion datasets: FER2013, RAF-DB, AffectNet and ExpW. ResEmoteNet was trained on each dataset using carefully selected hyperparameters, based on sensitivity analysis experiments. We evaluated batch sizes of 16, 32, 64, and 128, and found that a batch size of 16 consistently yielded superior performance, likely because more frequent weight updates facilitated faster convergence during training. Similarly, we experimented with epochs ranging from 40 to 80, with 80 epochs being the optimal choice, as the model’s performance saturated around this point, and further increases yielded minimal gains in accuracy. Thus, the final configuration employed a batch size of 16 and 80 training epochs across all datasets, using Cross-Entropy Loss \cite{zhang2018generalized} as the cost function and Stochastic Gradient Descent (SGD) \cite{ruder2016overview} as the optimizer. As the classes in all the datasets are not balanced as seen from Table \ref{classwise-data-stats}, we used a class-weight biased method for training, where the classes with fewer samples were given more weightage in the loss function. We applied Random Horizontal Flip for data augmentation and used a learning rate of $1 \times 10^{-3}$, with a scheduler that reduced the rate by 0.1 when performance plateaued. Our experiments were executed on two distinct hardware configurations: a MacBook Pro (M2 Pro-Chip with 10-core CPU and 16-core GPU) and a NVIDIA Tesla P100 GPU provided by Kaggle. The implementation was done using PyTorch \cite{paszke2019pytorch}. For our facial emotion recognition tasks, we employ Accuracy (\%) as the evaluation metric, defined as:

\begin{equation}
\label{eq:accuracy}
\text{Accuracy (\%)} = \frac{\text{TP} + \text{TN}}{\text{TP} + \text{TN} + \text{FP} + \text{FN}} \times 100
\end{equation}
where TP: True Positive, TN: True Negative, FP: False Positive, FN: False Negative. ResEmoteNet was trained for 3 hours for FER2013, 2 hours for RAF-DB, 6 hours for AffectNet-7 and 4.5 hours for ExpW, with inference times of 1.4ms, 3.2ms, 3.6ms and 1.9ms per image respectively.

\vspace{-0.25cm}
\section{Results and Discussion}\label{sec:results}

In this section, we present the experimental results conducted on four widely used benchmark datasets, FER2013, RAF-DB, AffectNet-7 and ExpW.
We evaluated our proposed method, ResEmoteNet, on these datasets and tabulated in Table \ref{result-compare}. We also present the confusion matrices of the ResEmoteNet for each of the datasets depicting their class wise confusions on the respective test sets in Fig. \ref{fig:confusion_matrices}.

\vspace{-0.25cm}
\subsection{Model Performance Across Datasets: Comparison with Prior Studies}
In Table \ref{result-compare}, we compared the performance of the proposed method with other state-of-the-art methods, the results demonstrate that our proposed method outperforms the current state-of-the-art techniques. 
Working with FER2013 is challenging due to inaccurate labeling, absence of faces in some images, and imbalanced data distribution, yet ResEmoteNet achieved a classification accuracy of 79.79\%, a 2.97\% absolute improvement over Ensemble ResMaskingNet \cite{residualmasking}. On RAF-DB, which presents real-life challenges such as pose, lighting, and occlusion, ResEmoteNet attained a classification accuracy of 94.76\%, improving by 2.19\% over S2D \cite{s2d}. On AffectNet-7, a large-scale dataset with 7 emotions and diverse annotations, our model achieved a classification accuracy of 72.93\%, showing a 3.53\% improvement over EmoAffectNet \cite{kollias2024distribution}. Lastly, on ExpW, which presents a broad range of facial expressions in uncontrolled, real world settings, ResEmoteNet achieved a classification accuracy of 75.67\%, showing a 2.19\% improvement over APViT \cite{xue2022vision}.

\renewcommand{\arraystretch}{1.3}
\begin{table}[!ht]
    \centering
    \small
    \caption{Test Accuracy (\%) comparison of ResEmoteNet with existing state-of-the-art methods across four datasets: FER2013, RAF-DB, AffectNet-7 and ExpW.}
    \scalebox{0.71}{
    \setlength{\tabcolsep}{10pt}
    \begin{tabular}{|c|c|c|c|c|}
        \hline
           & \multicolumn{4}{c|}{\bf {Accuracy in \%}} \\ 
 \cline{2-5}
            \textbf{Method} & \textbf{FER2013} & \textbf{RAF-DB} & \textbf{AffectNet-7} & \textbf{ExpW}\\
        \hline
         Seg. VGG-19 \cite{vignesh2023novel} & 75.97 & - & - & -\\
          \hline
         EmoNeXt \cite{el2023emonext} & 76.12 & - & - & -\\
          \hline
         En. ResMaskingNet \cite{residualmasking} & 76.82 & - & - & -\\
          \hline
         SEResNet \cite{huang2023study} & - &  83.37 & 56.54 & -\\
          \hline
         Arm \cite{shi2021learning} & - & 90.42 & 62.5 & -\\
         \hline
         APVit \cite{xue2022vision} & - & 91.98 & 66.91 & 73.48 \\
         \hline
         ARBEx \cite{arbex} & - & 92.47 & - & -\\
          \hline
         S2D \cite{s2d} & - & 92.57 & 67.62  & -\\
          \hline
         C MT EmoAffectNet \cite{kollias2024distribution} & - & - & 69.4 & -\\
         \hline
         AGLRLS \cite{gao2024adaptive} & - & - & - & 73 \\
         \hline
         SchiNet \cite{bishay2019schinet} & - & - & - & 73.10 \\
         \hline
         \textbf{Proposed ResEmoteNet} & \textbf{79.79} & \textbf{94.76} & \textbf{72.93} &\textbf{75.67} \\
         \hline
    \end{tabular}}
    \label{result-compare}
    \vspace{-1.2em}
\end{table}

\vspace{-0.25cm}
\subsection{Ablation Study}
We conducted an ablation study on the RAF-DB dataset to evaluate the contribution of key components: Convolutional layers, SE blocks, and Residual Blocks within the ResEmoteNet architecture. The results in Table \ref{tab:ablation_study} demonstrate the impact on accuracy and loss when these components are removed.

Removing the SE block resulted in a minor accuracy drop from 94.76\% to 94.18\%, while eliminating Residual Block 1 (with 256 channels) led to a substantial reduction in accuracy, from 94.76\% to 75.39\%, highlighting the critical role of residual connections. Similarly, removing convolutional layers, such as CNN 64 and CNN 64, 128, caused accuracy decreases of 11.25\% and 6.49\%, respectively.

\renewcommand{\arraystretch}{1.6}

\begin{table}[!ht]
\caption{Component-wise study of ResEmoteNet in RAF-DB Dataset. Note that Config C1 denotes our best ResEmoteNet; CNN stands for Convolutional Neural Network, SE stands for Squeeze-and-Excitation Block, RB stands for Residual Block respectively.}
\label{tab:ablation_study}
\centering
\scalebox{0.7}{
\begin{tabular}{|c|ccc|c|c|c|}
\hline
\multicolumn{1}{|l|}{\multirow{2}{*}{\textbf{Config}}} & \multicolumn{3}{c|}{\textbf{Components}}                                                                 & \multirow{2}{*}{\makecell{\textbf{Removed} \\ \textbf{Component(s)}}} & \multirow{2}{*}{\textbf{Accuracy (\%)}} & \multirow{2}{*}{\textbf{Loss (\%)}} \\ \cline{2-4}
\multicolumn{1}{|l|}{}                                        & \multicolumn{1}{c|}{\textbf{CNN}}          & \multicolumn{1}{c|}{\textbf{SE}}  & \textbf{RB}             &                                                &                                         &                                     \\ \hline
\textbf{C1}                                                   & \multicolumn{1}{c|}{\textbf{64, 128, 256}} & \multicolumn{1}{c|}{\textbf{256}} & \textbf{256, 512, 1024} & \textbf{-}                                     & \textbf{94.76}                          & \textbf{3.73}                       \\ \hline
C2                                                            & \multicolumn{1}{c|}{64, 128, 256}          & \multicolumn{1}{c|}{-}            & 256, 512, 1024          & SE (256)                                        & 94.18 (-0.58)                           & 16.83 (+13.1)                       \\ \hline
C3                                                            & \multicolumn{1}{c|}{128, 256}              & \multicolumn{1}{c|}{256}          & 256, 512, 1024          & CNN (64)                                       & 83.51 (-11.25)                          & 7.19 (+3.46)                        \\ \hline
C4                                                            & \multicolumn{1}{c|}{256}                   & \multicolumn{1}{c|}{256}          & 256, 512, 1024          & CNN (64, 128)                                  & 88.27 (-6.49)                           & 5.32 (+1.59)                        \\ \hline
C5                                                            & \multicolumn{1}{c|}{64, 128, 256}          & \multicolumn{1}{c|}{256}          & 512, 1024               & RB (256)                                       & 75.39 (-19.37)                          & 10.2 (+6.47)                        \\ \hline
C6                                                            & \multicolumn{1}{c|}{64, 128, 256}          & \multicolumn{1}{c|}{256}          & 1024                    & RB (256, 512)                                  & 82.77 (-11.99)                          & 8.8 (+5.07)                         \\ \hline
\end{tabular}}
\end{table}

\vspace{-0.5cm}
\section{Conclusion}\label{sec:conclusion}
In this paper, we introduced ResEmoteNet, a novel neural network architecture designed to address the challenging task of facial emotion recognition. Our model integrates a combination of three distinct networks: Convolutional Neural Network, Squeeze and Excitation network and residual network. The integration of these networks allows ResEmoteNet to effectively capture and interpret complex emotional expressions from facial images. To evaluate the performance of our proposed ResEmoteNet, we conducted extensive experiments using four widely recognized benchmark datasets: FER2013, RAF-DB, AffectNet-7 and ExpW. Our experimental results demonstrate that ResEmoteNet consistently outperforms existing state-of-the-art models across all these datasets. Additionally, an ablation study on the RAF-DB dataset confirmed the critical role of the residual and convolutional layers in maintaining the model’s performance. These results highlight the effectiveness of our approach in accurately recognizing facial emotions, offering significant advancements in the field of facial emotion recognition. Future work will explore further enhancements to ResEmoteNet and its applications in real-world scenarios.

\vspace{-0.5em}
\section{Acknowledgement}
This work was supported by the iHub DivyaSampark IIT Roorkee.

\bibliographystyle{ieeetr}
\bibliography{main}

\end{document}